\documentclass[conference]{IEEEtran}
\usepackage{times}
\usepackage[numbers]{natbib}
\usepackage{multicol}
\usepackage{booktabs}
\usepackage[bookmarks=true]{hyperref}
\usepackage{amsmath}
\usepackage{amssymb}
\usepackage{graphicx}
\usepackage{subcaption}

\usepackage{multirow}
\usepackage{multicol}
\usepackage{graphicx}
\usepackage{xspace}
\usepackage{xcolor}
\usepackage{caption}
\usepackage{wrapfig}
\usepackage{bbding}  
\usepackage{pifont}  

\usepackage{booktabs}
\usepackage{float}
\usepackage{amsmath}  
\usepackage{amssymb}  
\usepackage{bm}
\usepackage{bbm}
\usepackage{dsfont}
\usepackage{makecell}

\usepackage{hyperref}
\usepackage[capitalise, nameinlink]{cleveref}

\usepackage{colortbl}
\definecolor{ourcolor}{HTML}{99e0eb}
\definecolor{ourblue}{HTML}{27a2c3}

\definecolor{tablecolor}{HTML}{ccf2f5} 

\definecolor{tablecolor2}{HTML}{ffcdb4}
\definecolor{citecolor}{HTML}{fe7b5b}
\definecolor{grey}{rgb}{0.9, 0.9, 0.9}
\usepackage{amssymb}

\usepackage{listings}
\definecolor{gred}{rgb}{0.859,0.267,0.216}
\definecolor{ggreen}{rgb}{0.059,0.616,0.345}

\definecolor{deepblue}{HTML}{27a2c3}

\definecolor{deepred}{HTML}{fe7b5b}

\usepackage[font=small,labelfont=bf]{caption}

\usepackage[font=footnotesize,labelfont=bf]{caption}

\definecolor{citecolor}{HTML}{faa700} 
\definecolor{lblue}{HTML}{ffb114} 
\definecolor{ogreen}{HTML}{2E7D32}
\definecolor{bred}{HTML}{BF360C}
\definecolor{newbrown}{HTML}{795548}

\hypersetup{
    colorlinks=true,
    linkcolor=brown,
    filecolor=magenta,      
    urlcolor=brown,
    citecolor=brown,
}

\usepackage{multicol}
\usepackage{multirow}
\usepackage{colortbl}
\usepackage{booktabs}   
\usepackage{bbding} 
\usepackage{graphicx}
\let\labelindent\relax 
\usepackage{enumitem}

\usepackage{bbding}

\usepackage{pifont}
\usepackage{xfrac}
\usepackage{arydshln}

\begin{document}
\definecolor{myred}{RGB}{254,113,52}

\title{
A Scalable Whole-body Motion Transfer via \textcolor{myred}{I}mplicit \textcolor{myred}{K}inodynamic \textcolor{myred}{M}otion \textcolor{myred}{R}etargeting}
\author{
    Xingyu Chen\textsuperscript{1, 4}, 
    Hanyu Wu\textsuperscript{2, 4}, 
    Sikai Wu\textsuperscript{4}, 
    Mingliang Zhou\textsuperscript{4}, 
    Diyun Xiang\textsuperscript{4}, 
    Haodong Zhang\textsuperscript{3},\\
    Yangchen Zhou\textsuperscript{1},
    Yukang Gao\textsuperscript{1},
    Yi Gu\textsuperscript{1},
    Renjing Xu\textsuperscript{1}\\
\authorblockA{
\textsuperscript{1}HKUST(GZ)\quad 
\textsuperscript{2}ETH Zurich\quad 
\textsuperscript{3}Zhejiang University\quad 
\textsuperscript{4}Xiaomi Robotics Lab\\
\href{https://cybercal.github.io/webpage.ikmr/}{\texttt{https://cybercal.github.io/webpage.ikmr}}
\vspace{-4pt}
}
}

\twocolumn[{%
\renewcommand\twocolumn[1][]{#1}%
\maketitle

}]

\begin{abstract}

Human-to-humanoid imitation learning presents a promising pathway to address the severe data scarcity bottleneck in robotics by utilizing abundant, large-scale human motion collections. 
However, scaling this paradigm requires addressing two key challenges.
First, human motion data acquired from videos, motion capture systems, or generative models often contains spatial noise, jitter, and frame-level flickering, which can be amplified during retargeting and lead to unsafe or physically infeasible robot motions.
Second, existing motion retargeting methods typically rely on frame-by-frame numerical optimization, making them too computationally expensive for large-scale dataset synthesis.
To overcome these limitations, we introduce Implicit Kinodynamic Motion Retargeting (IKMR), a highly scalable, neural-based data transformation pipeline. 
IKMR leverages a skeleton-based graph convolutional dual autoencoder to map cross-structural human and humanoid kinematic configurations into a shared topological latent space. 
To guarantee the physical viability of the generated data, the framework incorporates a physics-informed refinement phase that utilizes simulated physical tracking feedback to learn a robust motion prior. 
This implicit formulation fundamentally resolves both challenges. By shifting the computational burden from online optimization to offline inference, IKMR achieves an unprecedented data conversion throughput exceeding 5000 frames per second. 
Furthermore, leveraging the learned motion prior, it functions as an intrinsic data curation mechanism and naturally filters out high-frequency noise and spatial jitters from source data, yielding smooth trajectories that ensure physical hardware safety. 
Extensive evaluations, including real-world whole-body control deployments on humanoid robot Unitree G1, confirm that IKMR successfully bridges the gap between unconstrained human motion dataset and robotic data.

\end{abstract}

\IEEEpeerreviewmaketitle

\section{Introduction}
\label{sec:intro}

\begin{figure}[t]
  \centering
    \includegraphics[width=1.0\linewidth]{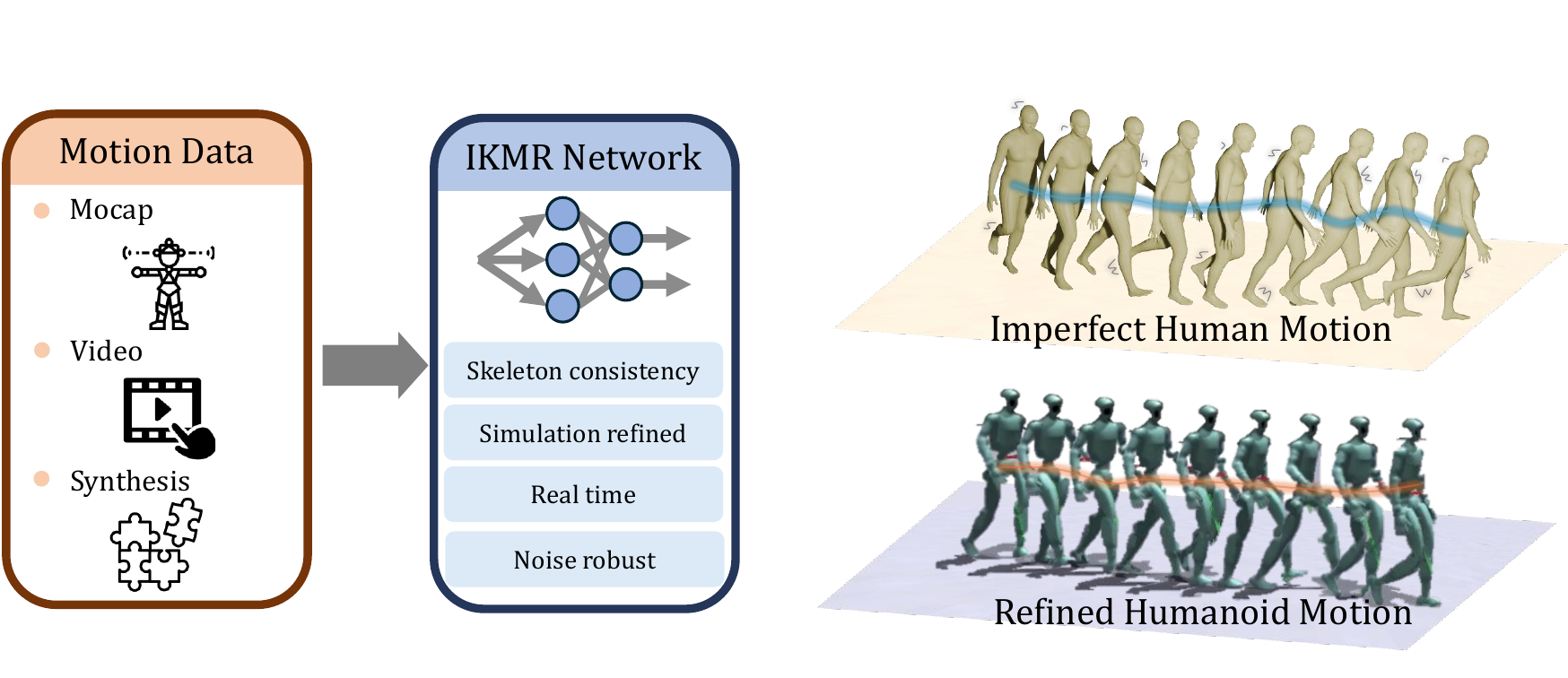}
    \caption{\textbf{IKMR as a scalable data transformation pipeline.}
    Abundant but imperfect human motion data—sourced from mocap, in-the-wild videos, or generative models—often contains spatial noise and jitters. Our fully vectorized IKMR framework ingests these heterogeneous datasets and implicitly filters out disturbances through physics-informed refinement, efficiently converting unconstrained human movements into robot-executable trajectories.
    }
    \vspace{-11pt}
    \label{fig:task}
\end{figure}

Acquiring diverse and agile motor repertoires on humanoid robots remains a central challenge in whole-body imitation learning~\cite{ze2025twist, he2024omnih2o, cheng2024expressive, xue2025unified, allshire2025visual, yin2025unitracker, chen2025gmt}. Unlike natural language processing and computer vision, where large-scale datasets are readily available, physical robotics is constrained by the scarcity of diverse, high-quality robot motion data. Directly collecting extensive and high-fidelity locomotion trajectories on physical humanoid hardware is profoundly impractical due to prohibitive operational costs and the risk of mechanical degradation~\cite{rotella2014state, kuindersma2016optimization}. Furthermore, physical data collection is an inherently open-ended endeavor. A sustainable data-driven paradigm must transcend the brute-force accumulation of disjointed long-tail corner cases. Instead, it should establish a continuous assimilation loop—where acquiring foundational data structurally bootstraps the robust ingestion and generalization of subsequent, more complex experiences. In contrast, human motion data is exceptionally abundant, highly diverse, and expanding at an unprecedented scale~\cite{lin2023motion, wang2024scaling}. High-quality human movements can be acquired through optical motion capture systems~\cite{kirk2004skeletal, mundermann2006evolution, menolotto2020motion}, extracted directly from in-the-wild internet videos~\cite{wang2024tram, shen2024world, shin2024wham, muller2025reconstructing}, or synthesized via generative diffusion and autoregressive models conditioned on multi-modal inputs~\cite{tang2018dance, plappert2016kit, guo2022generating, punnakkal2021babel, bhatnagar2022behave, wang2022humanise, tevet2023human, zhang2022motiondiffuse, chen2023mld, chen2024text, ghosh2021synthesis, tevet2022motionclip, zhang2023t2m, guo2024momask}. This colossal reservoir of cross-structural human data serves as the ideal catalyst for this continuous loop, holding immense potential to compensate for the scarcity of robotic training assets, as depicted in Figure~\ref{fig:task}. Therefore, from a data-centric perspective, a fundamental challenge remains unresolved: \textit{How can heterogeneous, unconstrained human motion datasets be efficiently transformed into high-fidelity, robot-executable training data at scale?}

The primary barrier to realizing this data-driven paradigm resides within the motion transformation layer. Specifically, translating massive human datasets into robotic trajectories requires addressing two critical challenges. 
First, the issue of \textit{computational scalability}. To transfer trajectories from a human skeleton to a humanoid structure with distinct joint limits and links, conventional pipelines heavily rely on explicit numerical optimization applied frame-by-frame~\cite{choi2000online, monzani2000using, darvish2019whole, schumacher2021versatile, luo2023perpetual}. While these kinematics-based approaches achieve geometric alignment, they suffer from prohibitive computational overhead. This frame-level reliance completely lacks the computational tractability required for batch processing and large-scale dataset synthesis. 
Second, the issue of \textit{data quality and hardware safety}. Real-world human data harvested from videos, motion capture systems, or generative networks is frequently plagued by spatial noise, jitter, and frame-level flickering. Explicit numerical methods tend to blindly overfit to these spatial perturbations, amplifying the artifacts into high-frequency joint tremors~\cite{tosun2014general, gleicher1998retargetting}. Executing such trajectories on physical hardware severely compromises dynamic feasibility and leads to highly unsafe robot behaviors. Although dynamics-based methods incorporate manual physical constraints~\cite{popovic1999physically}, they exacerbate the computational cost. Existing neural implicit retargeting approaches improve efficiency~\cite{choi2020nonparametric, villegas2018neural, aberman2020skeleton, zhao2024pose}, but predominantly focus on virtual avatars and ignore dynamic constraints, rendering the generated motions physically infeasible for actual hardware. 

To simultaneously address these two challenges and bridge the data-to-control gap, we propose \textbf{Implicit Kinodynamic Motion Retargeting (IKMR)}, a framework that reformulates traditional motion conversion into a high-throughput, noise-robust data generation pipeline for humanoid learning. Our core insight is that although humans and humanoid robots exhibit distinct physical geometries, they possess a homeomorphic and topologically equivalent kinematic chain. Utilizing a skeleton-based graph convolutional dual autoencoder, IKMR projects cross-structural configurations into an aligned, topology-aware latent representation space. To ensure the generated data strictly satisfies physical constraints, we embed a physics-informed simulation tracking loop into the post-training phase to fine-tune the decoder. This design shifts the heavy computational burden of ensuring dynamic feasibility from active, run-time optimization to passive, offline network inference. By transitioning from per-frame optimization to fully vectorized neural mapping, IKMR fundamentally learns a robust motion prior distribution from the training data. This structural prior endows the framework with strong generalization capabilities against noisy or degraded source motions. Consequently, IKMR acts as the crucial engine for the aforementioned continuous assimilation loop: existing learned priors ensure that newly encountered unconstrained motions are safely regularized, preventing the explosion of edge cases and allowing the robot to smoothly and endlessly decode dynamically feasible trajectories.

The principal contributions are summarized as follows: 

\begin{itemize}
    \item \textbf{Scalable human-to-humanoid motion transformation.}
    We introduce IKMR, a fully vectorized neural retargeting framework that maps heterogeneous skeleton into a shared topological latent space, enabling high-throughput dataset-level motion conversion.

    \item \textbf{Physics-informed motion refinement.}
    We propose a simulator-in-the-loop strategy that improves physical feasibility while implicitly suppressing noise, jitter, and frame-level flickering in unconstrained source motions.

    \item \textbf{Real-world humanoid validation.}
    We validate IKMR through retargeting comparison experiments and downstream whole-body control deployment on the Unitree G1, demonstrating its ability to convert human motion into robot-executable data.
\end{itemize}
\section{Related Works}
\label{sec:rela}

\begin{table}[t]
    \centering
    \resizebox{\linewidth}{!}{
        \begin{tabular}{lcccc}
            \toprule
            \textbf{Metric}
            & \makecell{\textbf{PHC}~ \cite{luo2023perpetual}}
            & \makecell{\textbf{Mink}~ \cite{zakka10mink}}
            & \makecell{\textbf{GMR}~ \cite{joao2025gmr}}
            & \makecell{\textbf{IKMR}~ \textbf{(Ours)}} \\
            \midrule
            Retargeting paradigm
            & Optimization
            & Optimization
            & Optimization
            & Neural inference \\
            Kinematic formulation
            & FK
            & IK
            & IK
            & Implicit \\
            Throughput
            & 21 FPS
            & 49 FPS
            & 64 FPS
            & \textbf{5000 FPS} \\
            Real-time capable
            & $\times$
            & $\checkmark$
            & $\checkmark$
            & $\checkmark$ \\
            Batch scalable
            & $\times$
            & $\times$
            & $\times$
            & $\checkmark$ \\
            \bottomrule
        \end{tabular}
    }
    \caption{
    Comparison with representative retargeting methods for whole-body humanoid control.
    Runtime is measured on an Intel Core i7-14700HX CPU and an NVIDIA RTX 4090 GPU.
    Batch scalability indicates whether the method supports vectorized parallel inference for large-scale motion dataset synthesis.
    }
    \label{table:1}
    \vspace{-5pt}
\end{table}

\subsection{Human Motion Retargeting}
Early motion retargeting methods primarily focused on optimization with kinematic objectives and constraints~\cite{tosun2014general, gleicher1998retargetting}, treating it as a fitting problem of joint positions or link orientation between source and target skeletons.
Then, the inverse kinematics method~\cite{lee1999hierarchical} reserves the end-effector positions and computes the joint angles to achieve higher precision. To preserve essential physical properties of the motion, some research proposes spacetime constraints dynamics~\cite{popovic1999physically} to maintain realism of the original motion. An intermediate skeleton~\cite{monzani2000using} was explored to convert movements between hierarchically and geometrically different characters. For human-to-humanoid motion retargeting, current methods mainly employ a hierarchical process~\cite{luo2023perpetual, zakka10mink, joao2025gmr}, first matching body scales, and then optimizing joint poses. These approaches often operate frame-by-frame, and require optimizing parameters for each trajectory.

\subsection{Neural Motion Processing}
With the development of deep learning, the neural network performs a low-cost and high-efficiency approach to learn the motion mapping. The primary learning-based retargeting work~\cite{holden2016deep} employs a convolutional neural network to reduce the temporal dimension of the motion sequence, while it ignores the topology structure of the character skeleton. To address this problem, researchers begin to model the character skeleton as a graph~\cite{yan2018spatial}, where nodes and edges express joints and bones, respectively. Then, the convolutional kernel could be applied on the joint and its neighbors~\cite{ci2019optimizing}. Inspired by the CycleGAN~\cite{zhu2017unpaired}, researchers begin to introduce this architecture in the motion retargeting task, and develop it to an unsupervised paradigm~\cite{villegas2018neural, aberman2020skeleton}. Given the target skeleton, pose2motion~\cite{zhao2024pose} learn a pose prior that transfers source motion without target motion in the training dataset. Recently, a dynamic graph transformation module~\cite{zhang2024unified} is proposed to handle unseen skeletal structures for humanoid motion retargeting.

\begin{figure*}[ht]
  \centering
  \includegraphics[width = 1.0\textwidth]{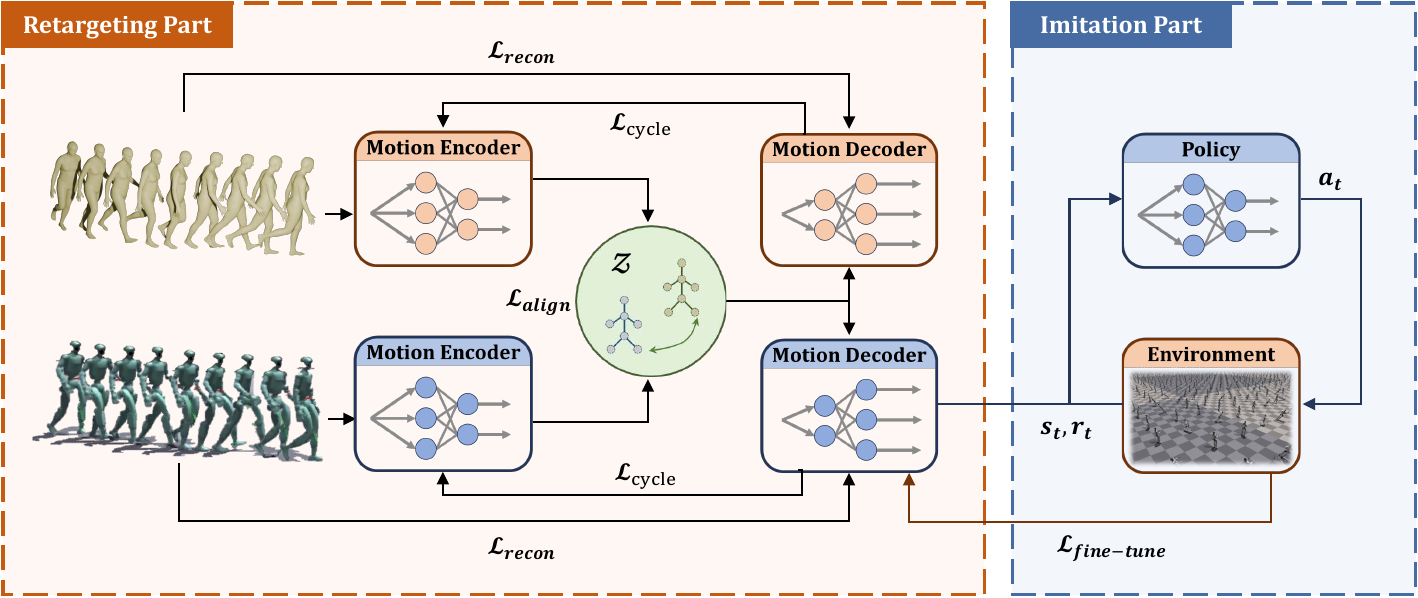}
    \caption{\textbf{Overview of the IKMR data transformation pipeline.} The framework bridges the domain gap between unconstrained human motion and physically feasible robotic trajectories through two core phases. Kinematics-Aware Pretraining: A dual encoder-decoder architecture projects heterogeneous human and humanoid motions into a shared topological latent space ($\mathcal{Z}$) via alignment ($\mathcal{L}_{align}$) and reconstruction ($\mathcal{L}_{recon}$) and cycle objectives ($\mathcal{L}_{cycle}$). Physics-informed Finetuning: A simulator-in-the-loop imitation policy interacts with the environment and provides physical feedback ($\mathcal{L}_{fine-tune}$). By fine-tuning the target motion decoder with this dynamic feedback, IKMR learns a robust motion prior, enabling the pipeline to automatically purify imperfect source motion and decode as robotic data.}
  \label{fig:framework}
  \vspace{-10pt}
\end{figure*}

\subsection{Humanoid Imitation Learning}
DeepMimic~\cite{peng2018deepmimic} pioneers imitation learning for humanoid characters by deep reinforcement learning. It directly applies the tracking error as a reward to encourage the agent to fit the trajectory. Then ASAP~\cite{he2025asap} proposed a delta action model to reduce the sim to real gap between simulator and real robot. KungfuBot~\cite{xie2025kungfubot} introduce an adaptive sigma for each reward function for high precision tracking. Also, some researches model the motion transition as a distribution, AMP~\cite{peng2021amp} applies an adversarial network to provide a style reward for imitating the state transition. Early imitation learning methods usually train a policy to track one reference motion, ASE~\cite{peng2022ase} develops a high-level controller for scheduling fundamental skills. To adapt for tele-operation and versatile locomotion, more advanced imitation methods have recently been proposed, such as OmniH2O~\cite{he2024omnih2o}, Exbody~\cite{cheng2024expressive}, HumanPlus~\cite{fu2024humanplus}, TWIST~\cite{ze2025twist} and GMR~\cite{joao2025gmr}. Furthermore, humanoid foundation models~\cite{luo2025sonic, zeng2025behavior, yuan2025behavior} strictly rely on extreme-fidelity data. We introduce IKMR as a scalable domain-alignment pipeline that safely projects noisy, unconstrained imperfect motions into a dynamically feasible representation space, effectively unlocking the vast potential of in-the-wild data without triggering out-of-distribution degradation.

\section{Proposed Method}

\subsection{Problem Formulation}

In general, kinematic motion representations comprise two essential components: a static structure detailing the physical embodiment, and a dynamic sequence expressing state transitions over time. For the static branch, the representation relies on a relational skeleton tree and an initial rest pose. The skeleton tree illustrates the hierarchical structure of the embodiment, specifying the connections and kinematic constraints between joints and links via nodes and edges. We define the static skeleton as $S \in \mathbb{R}^{J \times 3}$, where $J$ denotes the number of joints, and the 3D vectors represent the static joint offsets in the initial local coordinate frame. 

For the dynamic branch, the motion sequence comprises root translations and local joint rotations. The root translation describes the global trajectory of the root node over a temporal window. The joint rotations are parameterized as $Q \in \mathbb{R}^{T \times J \times 4}$, describing the local rotational state of each joint relative to its parent frame across $T$ time steps using unit quaternions. 

The problem of data-driven motion transformation is formalized as follows. Let $M_A = (S_A, Q_A)$ denote the source human motion and $M_B = (S_B, Q_B)$ denote the target humanoid robot motion. We define a cross-structural training dataset $\mathcal{D} = \{(M^i_A, M^i_B)\}_{i=1}^N$, where each tuple comprises paired human and robot motion sequences. To overcome the computational bottlenecks of frame-by-frame numerical optimization, our implicit transformation pipeline aims to learn a fully vectorized functional mapping that directly translates the source dynamics $Q_A$ into the target dynamics $Q_B$ conditioned on their respective skeletal structures, mathematically denoted as $\mathit{f} : (Q_A, S_A, S_B) \rightarrow Q_B$. 

\subsection{Topology-Aware Motion Representation}

\begin{figure}[h]
  \centering
    \includegraphics[width=0.9\linewidth]{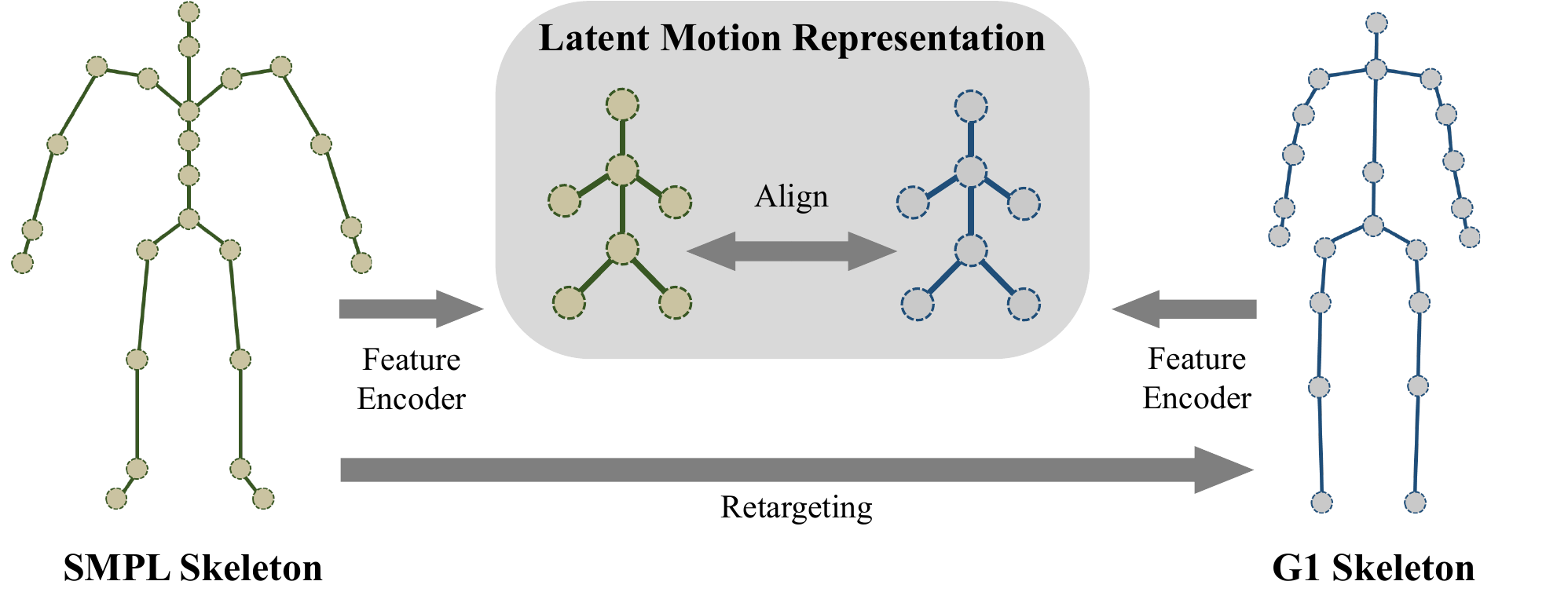}
    \caption{Latent motion feature alignment between the human and humanoid robot. IKMR encodes motions from heterogeneous structures into a shared, topology-aware representation space.}
    \label{fig:topoalign}
    \vspace{-10pt}
\end{figure}

For human-to-humanoid motion transformation, a foundational insight is that the source skeleton $S_A$ and the target skeleton $S_B$ are homeomorphic. Despite significant discrepancies in bone lengths, degrees of freedom, and mechanical joint limits, they share a topologically equivalent kinematic chain (e.g., all limbs ultimately route to a central torso). Directly fitting explicit joint angles or end-effector positions across such heterogeneous structures inevitably introduces severe geometric bias and kinematic artifacts.

To construct a scalable domain-alignment pipeline, we abandon explicit spatial fitting. Instead, we enforce consistency within a shared latent feature space. Specifically, regardless of the structural differences in $S_A$ and $S_B$, their temporal dynamics can be projected into a unified set of topological prototypes, as illustrated in Figure~\ref{fig:topoalign}. By establishing this shared representation space, IKMR fundamentally decouples the motion content from its physical embodiment, enabling high-throughput, sequence-level data conversion without the overhead of per-frame inverse kinematics calculations.

\subsection{Kinematics-Aware Pretraining}

\begin{figure}[h]
  \centering
    \includegraphics[width=0.85\linewidth]{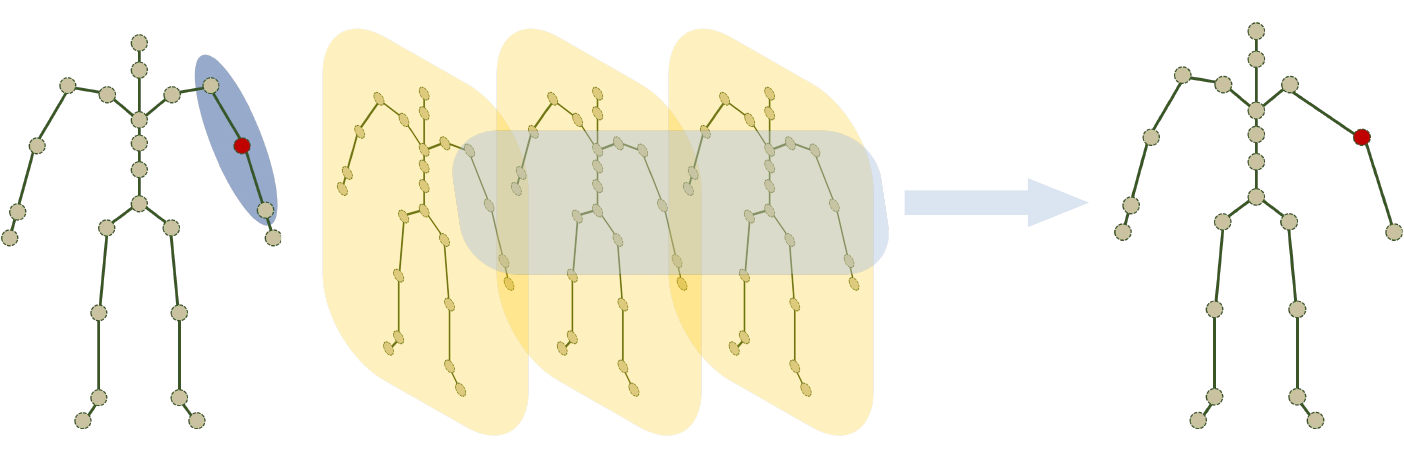}
    \caption{Skeleton-based graph convolution and pooling. The motion encoder extracts spatial-temporal latent features for each joint and its topological neighbors.}
    \label{fig:arch1}
    \vspace{-8pt}
\end{figure}

To effectively extract spatial-temporal latent features constrained by the skeletal topology, IKMR employs skeleton-based graph convolutional and pooling layers, as depicted in Figure~\ref{fig:arch1}. For each joint $i$, its hidden feature representation $\hat{M}_i$ is computed by aggregating information from its topological neighborhood $\mathcal{N}_i$:
\begin{equation} 
    \hat{M}_i = \frac{1}{|\mathcal{N}_i|} \sum_{j \in \mathcal{N}_i} \left( (Q_{j} \oplus S_{j}) \ast W_{j} + b_j \right)
\end{equation}
where $\oplus$ denotes feature concatenation, and the convolutional kernel is parameterized by weights $W_j$ and biases $b_j$. The graph convolution traverses the entire kinematic tree, ensuring that all topologically adjacent joints contribute to the local spatial-temporal receptive field. Subsequently, an average pooling operation condenses the structural features:
\begin{equation} 
    \text{Pooling} \{\hat{M}_i \} = \frac{1}{|\mathcal{N}_i|} \sum_{j \in \mathcal{N}_i}{\hat{M}_j}
\end{equation}

\begin{figure}[h]
  \centering
    \includegraphics[width=0.85\linewidth]{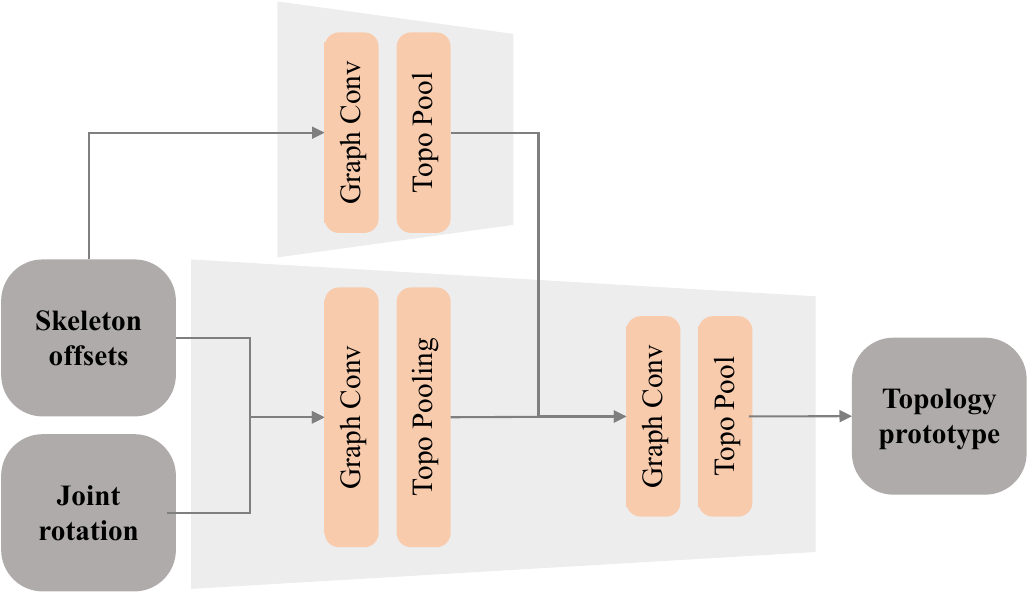}
    \caption{Layer structure in the motion encoder. By integrating the static and dynamic branches via graph convolution and pooling, source configurations are compressed into a unified topological motion representation.}
    \label{fig:arch2}
\end{figure}

For the transformation pipeline, we adopt a symmetric dual autoencoder architecture. As shown in Figure~\ref{fig:arch2}, each encoder-decoder pair comprises a static structural branch and a dynamic motion branch. The dynamic branch concatenates the skeleton offsets $S$ with the rotational sequences $Q$ prior to executing the graph convolutions.

Let $\mathcal{Z}$ denote the shared latent space. We define the source and target encoders as $\mathbf{E}_\theta^A$ and $\mathbf{E}_\theta^B$, mapping inputs to latents $z_A = \mathbf{E}_\theta^A(Q_A, S_A)$ and $z_B = \mathbf{E}_\theta^B(Q_B, S_B)$. Symmetrically, the decoders $\mathbf{D}_\phi^A$ and $\mathbf{D}_\phi^B$ reconstruct the motions from the latent space. The core cross-domain transformation is thus executed seamlessly as $\hat{Q}_B = \mathbf{D}_\phi^B(\mathbf{E}_\theta^A(Q_A, S_A), S_B)$. 

During the kinematics-aware pretraining stage, the network is optimized via three primary objectives. First, the \textit{Topology alignment loss} serves as the core regularization term, coercing the source and target encoders to project corresponding paired motions into identical latent coordinates:
\begin{equation} 
    \mathcal{L}_{align} = \mathbb{E}_{(M_A, M_B) \sim \mathcal{D}} \left\| \mathbf{E}_\theta^A(Q_A, S_A) - \mathbf{E}_\theta^B(Q_B, S_B) \right\|^2
\end{equation}
Second, the \textit{Reconstruction loss} guarantees that the compressed embeddings retain sufficient high-frequency detail to identically recover the original trajectories within their respective domains:
\begin{equation}
    \begin{aligned}
        \mathcal{L}_{recon} &= \mathbb{E}_{M_A \sim \mathcal{D}} \left\| \mathbf{D}_\phi^A(z_A, S_A) - Q_A \right\|^2 \\
        &+ \mathbb{E}_{M_B \sim \mathcal{D}} \left\| \mathbf{D}_\phi^B(z_B, S_B) - Q_B \right\|^2
    \end{aligned}
\end{equation}
Third, the \textit{Cycle consistency loss} ensures that motions mapped across domains and re-encoded maintain semantic stability, preventing latent space degradation:
\begin{equation} 
    \mathcal{L}_{cycle} = \mathbb{E}_{M_A \sim \mathcal{D}} \left\| \mathbf{E}_\theta^A(\mathbf{D}_\phi^A(z_A, S_A), S_A) - z_A \right\|^2
\end{equation}
The total objective function for the pretraining stage is formulated as:
\begin{equation} 
    \mathcal{L}_{pretrain} = \mathcal{L}_{recon} + \lambda_{align} \mathcal{L}_{align} + \lambda_{cycle}\mathcal{L}_{cycle} 
\end{equation}

\subsection{Physics-Informed Finetuning}

While kinematics-aware pretraining successfully aligns the topological space, the outputs may still contain spatial noise or violate strict hardware constraints. To endow IKMR with a robust motion prior and transform unconstrained data into dynamically feasible trajectories, we introduce a simulator-in-the-loop finetuning phase.

We formulate this phase using a standard reinforcement learning paradigm, where a tracking policy $\pi_{\theta}$ interacts with a physics simulator to maximize the cumulative expected reward $J(\theta)$:
\begin{equation} 
    J(\theta) = \mathbb{E}_{\tau \sim \pi_{\theta}(\tau)} \left[ \sum_{t=0}^{T} \gamma^t r_t \right]
\end{equation}
By training policies to track the pretrained reference motions across versatile skills (e.g., walking, running, and upper-body manipulations), the environment acts as a strict physical filter. The primary reward function encourages precise joint position tracking:
\begin{equation} 
    r_t =\exp\left(- \frac{\|q_t - \hat{q}_t\|_2^2}{\sigma_{j\text{pos}}}\right)
\end{equation}
where $q_t$ is the reference target position and $\hat{q}_t$ is the dynamically simulated state. Additional penalties for joint limits, collisions, and excessive torques guarantee stability. 

Once the policy converges, the simulator naturally curates a highly purified, physically feasible dataset. Utilizing these dynamically verified trajectories, we execute the final finetuning step. We freeze the source encoder $\mathbf{E}_\theta^A$ and strictly finetune the target decoder $\mathbf{D}_\phi^B$. To guarantee precise spatial interactions, we incorporate a Forward Kinematics (FK) end-effector loss, which enforces accuracy at critical contact points (hands and feet):
\begin{equation} 
    \mathcal{L}_{ee} = \mathbb{E} \left[ \left\| \text{FK}(\mathbf{D}_\phi^B(z_A, S_B)) - \text{FK}(\hat{Q}_B) \right\|^2 \right]
\end{equation}
The final objective function for physics-informed finetuning is thus given by:
\begin{equation} 
    \mathcal{L}_{finetune} = \mathcal{L}_{recon} + \lambda_{ee} \mathcal{L}_{ee} 
\end{equation}

By absorbing the simulator's feedback during finetuning, IKMR fundamentally learns a hardware-safe motion prior. This allows the pipeline to act as an intrinsic data curation mechanism, effortlessly filtering out in-the-wild noise and translating imperfect human motions into highly refined robotic datasets.
\section{Experiment}

During kinematics-aware pretraining, our model is trained on 2,200 human-to-humanoid trajectory pairs. The source human motions are sampled from the AMASS dataset~\cite{mahmood2019amass}, while the paired target motions are initially generated via a rudimentary kinematic joint mapping. For physics-informed finetuning, we train six distinct whole-body control policies within a physics simulator to sample 440 physically feasible motion trajectories, encompassing diverse skills such as walking, running, kicking, and upper-body manipulations. To evaluate the downstream efficacy of our data transformation pipeline, we conduct extensive experiments in both simulation and the real world using the compact humanoid robot Unitree G1 (132 cm height, 35 kg weight, 29 degrees of freedom).

\subsection{Topology Feature Analysis}
\begin{figure}[h]
  \centering
    \includegraphics[width=1.0\linewidth]{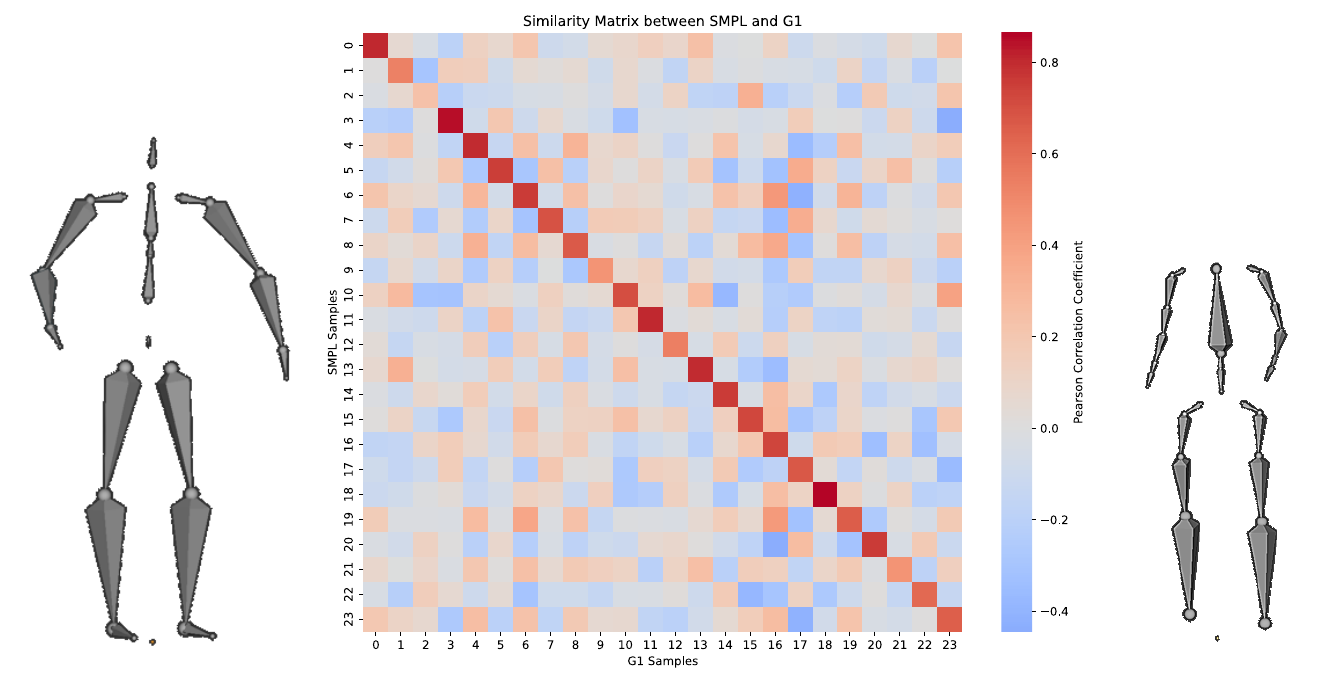}
    \caption{Correlation matrix of topological latent features. We sample 24 paired trajectories representing distinct motion types and extract their latent representations via the motion encoders. Feature similarity is quantified using the Pearson correlation coefficient.}
    \label{fig:unify}
    \vspace{-4pt}
\end{figure}

To quantitatively evaluate the efficacy of our unified topological representation, we select 24 distinct paired motion clips and utilize the Pearson correlation coefficient to measure the similarity between the encoded latent vectors of the human and humanoid motions. As illustrated in Figure~\ref{fig:unify}, our framework achieves strong topological feature alignment across disparate physical embodiments. In the similarity matrix, the maximum correlation coefficient reaches $0.8638$. This value indicates that while feature alignment is highly effective, it remains inherently lossy—a mathematically necessary outcome to accommodate the stark structural disparities between humans and robots. Furthermore, the mean correlation coefficient is $0.0195$, closely approximating zero. This demonstrates that the latent space exhibits a uniformly distributed and well-disentangled structure, devoid of collapse or over-clustering in any specific region. Consequently, our topological representation successfully preserves unique motion patterns, yielding high similarity for paired cross-structural motions and low similarity for unpaired ones.

\subsection{Physics-Informed Finetuning}

\begin{table}[h]
    \centering
    \begin{tabular}{lccc}
    \toprule
    \textbf{Phase}  &  \textbf{Mean Acc} $\downarrow$ & \textbf{Mean Jerk} $\downarrow$  \\
    \midrule
    
    Pretrain
    & 30.20 $\mathrm{rad/s^2}$
    & 456.60 $\mathrm{rad/s^3}$\\

    Finetune 
    & \textbf{29.15} $\mathrm{rad/s^2}$
    & \textbf{435.37} $\mathrm{rad/s^3}$\\
    \bottomrule
    \end{tabular}
    \caption{Comparison of trajectory smoothness. Physics-informed finetuning significantly mitigates high-frequency acceleration and jerk, producing hardware-safe robotic data.}
    \label{table:2}
\end{table}

While purely kinematics-based conversion achieves low static positional errors, the resulting motions frequently exhibit physically unrealistic artifacts, such as foot sliding, self-penetration, and floating. Our physics-informed finetuning phase effectively mitigates these discrepancies by embedding a hardware-safe motion prior into the decoder, yielding smooth and dynamically feasible trajectories.

\begin{figure*}[ht]
  \centering
    \includegraphics[width=1.0\linewidth]{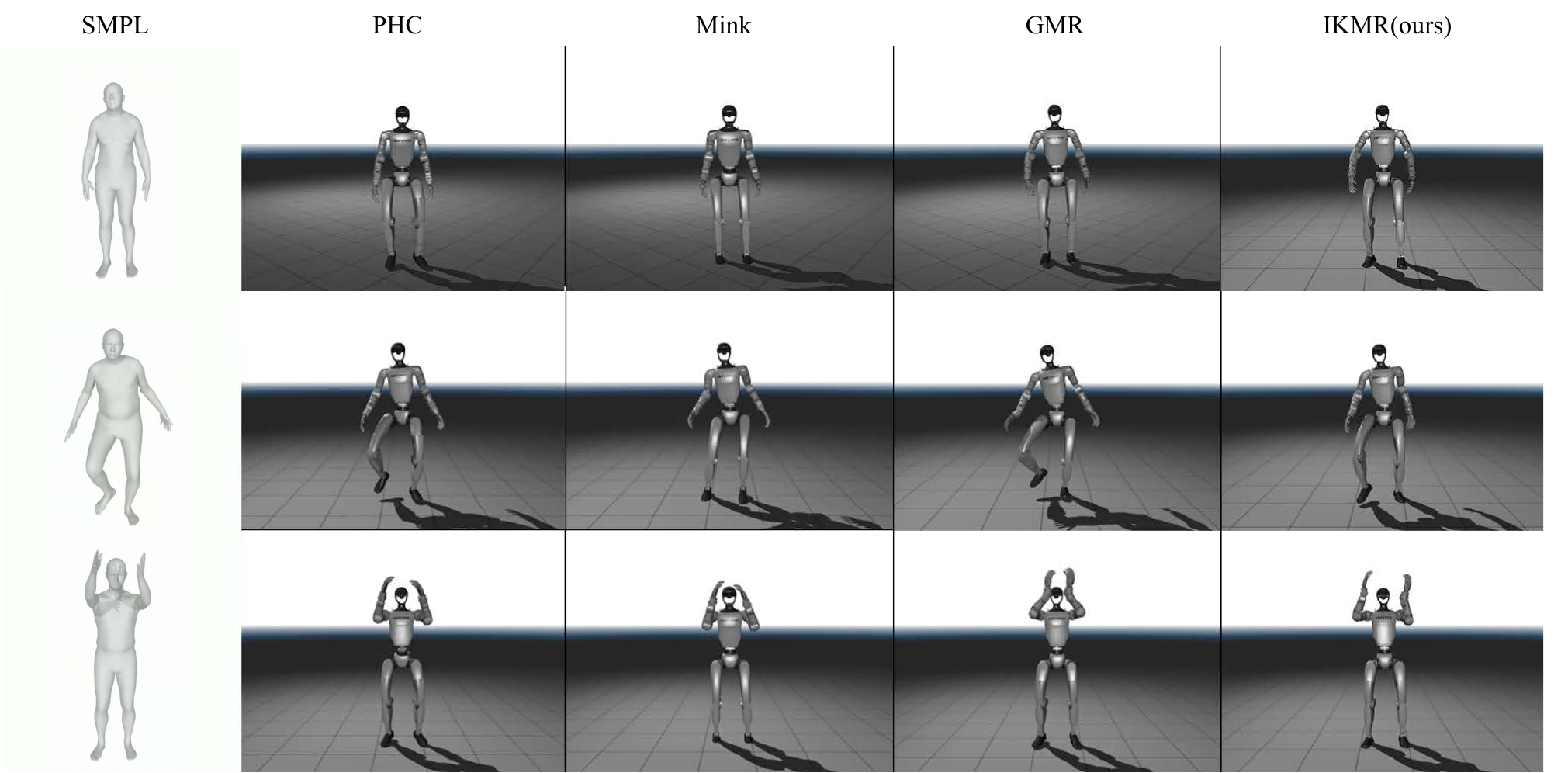}
    \caption{Visualization of key frames from common motion transfer methods: PHC~\cite{luo2023perpetual}, Mink~\cite{zakka10mink}, GMR~\cite{joao2025gmr}, and IKMR (Ours). The leftmost column displays the source human motion in SMPL format~\cite{loper2023smpl}, while the subsequent columns show the transformed motions for the Unitree G1 robot. IKMR successfully generates diverse, physically feasible trajectories through its implicit feature encoding-decoding pipeline.}
    \label{fig:perform}
    \vspace{-3pt}
\end{figure*}

To quantitatively validate this, we introduce joint acceleration and jerk as evaluation metrics to compare the transformed trajectories before and after finetuning. Acceleration measures the rate of change of angular velocity, while jerk measures the rate of change of acceleration, directly reflecting the smoothness and physical coordination of the generated movement.

We randomly sample 2,400 corresponding motion clips from both the pretrained and finetuned outputs to calculate the average per-joint acceleration and jerk. The comparison results, summarized in Table~\ref{table:2}, demonstrate that given the identical in-the-wild human reference trajectory, the physics-informed finetuned model outputs significantly lower acceleration and jerk values. By absorbing the simulator's dynamic constraints, the decoder effectively learns a robust motion prior tailored to the topology of the target robot. Consequently, the finetuned trajectories exhibit substantially smoother and more coordinated execution, ensuring physical hardware safety while faithfully preserving the semantic characteristics of the source data.

\subsection{Comparison and Robustness Analysis}

We visualize key frames of the transformed trajectories in Figure~\ref{fig:perform} and Figure~\ref{fig:visresults}, comparing our IKMR pipeline with three prevalent numerical whole-body control methods: PHC~\cite{luo2023perpetual}, Mink~\cite{zakka10mink}, and GMR~\cite{joao2025gmr}. In terms of computational efficiency, tested on an Intel Core i7-14700HX and RTX 4090, our method achieves a remarkable data transformation throughput of 5,000 FPS—nearly 100 times faster than frame-by-frame numerical optimization techniques (as detailed in Table~\ref{table:1}). Crucially, IKMR supports fully vectorized tensor operations, enabling rapid batch processing for massive dataset-level synthesis.

Beyond computational scalability, a vital capability for processing in-the-wild datasets is robustness against source data noise (e.g., measurement jitter in video motion capture). By implicitly mapping inputs into a topological latent space, IKMR functions as an intrinsic data curation mechanism. 

To evaluate this robustness, we artificially inject zero-mean Gaussian noise with increasing standard deviations into the root translations of the source human motions. We evaluate the outputs using two metrics:
\begin{itemize}
    \item \textit{Absolute Keypoint Trajectory Error (AKTE):} Adapted from visual SLAM evaluation, AKTE measures the absolute spatial deviation between the pristine target trajectory (derived from clean human motion) and the trajectory transformed from the noisy source. Evaluated on key joints (root, ankle, wrist, and knee), a lower AKTE denotes higher spatial consistency and resistance to source perturbations.

    \item \textit{Mean Key Joint Acceleration (MKJA):} This metric calculates the average linear acceleration of the aforementioned key joints. A lower MKJA signifies smoother, more coordinated, and hardware-safe robotic movement.
    
\end{itemize}

\begin{figure}[h]
  \centering
    \includegraphics[width=1.0\linewidth]{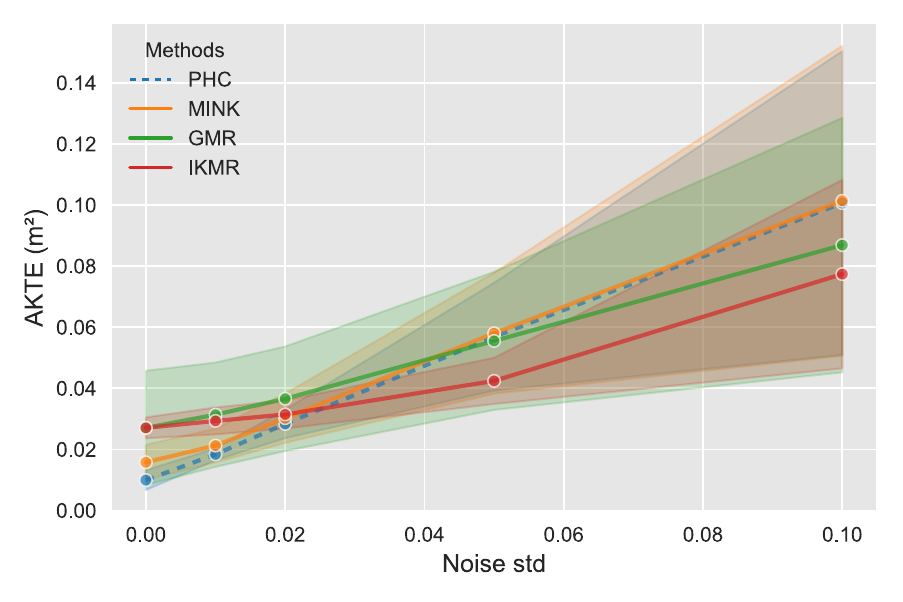}
    \includegraphics[width=1.0\linewidth]{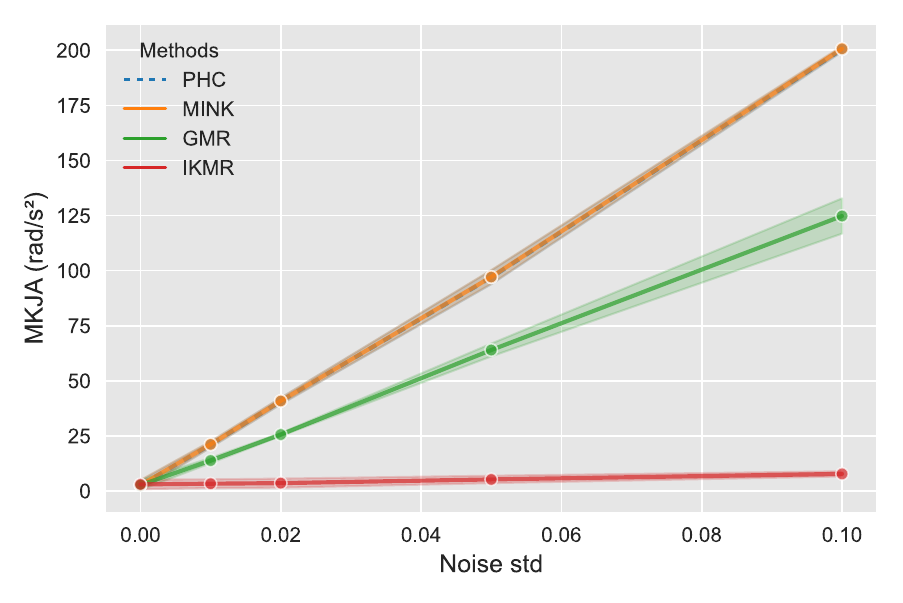}
\caption{\textbf{Quantitative evaluation of motion robustness under escalating source noise.} \textbf{(a) Absolute Keypoint Trajectory Error (AKTE):} As input noise increases, conventional explicit methods (PHC, MINK) exhibit severe spatial drift and high variance. In contrast, IKMR strictly bounds spatial errors, maintaining superior geometric consistency even at extreme noise levels. \textbf{(b) Mean Key Joint Acceleration (MKJA):} Explicit methods blindly overfit to spatial perturbations, resulting in a catastrophic explosion of high-frequency joint tremors (approaching 200 $\mathrm{rad/s^2}$). Conversely, IKMR maintains a near-invariant, low acceleration profile across all noise scales. Together, these results validate IKMR's capacity to act as a powerful intrinsic data filter, seamlessly purifying noisy, in-the-wild inputs into smooth, hardware-safe robotic trajectories.}
    \label{fig:line}
    \vspace{-6pt}
\end{figure}

As depicted in Figure~\ref{fig:line}, when the noise level escalates, explicit numerical methods indiscriminately fit the spatial perturbations, yielding sudden, aggressive local accelerations (high MKJA) and substantial trajectory drifts. 

In contrast, IKMR securely regularizes the unconstrained inputs, maintaining significantly lower and stable metric values. This demonstrates that our pipeline automatically purifies local tremors. This profound tolerance for noise significantly enhances motor safety, establishing IKMR as an indispensable, high-fidelity data bridge for scaling humanoid foundation models using imperfect real-world captures.

\subsection{Real-World Hardware Validation}

To validate the downstream utility of our transformed datasets, we deploy the generated trajectories on the physical Unitree G1 robot. By feeding the IKMR-curated robotic data into a standard reinforcement learning tracking pipeline, whole-body imitation controllers are directly trained and subsequently deployed on hardware. 

We evaluate the framework across diverse motor skills encompassing multi-directional and full-body coordination:

\begin{figure*}[ht]
  \centering
  \includegraphics[width = 0.99\textwidth]{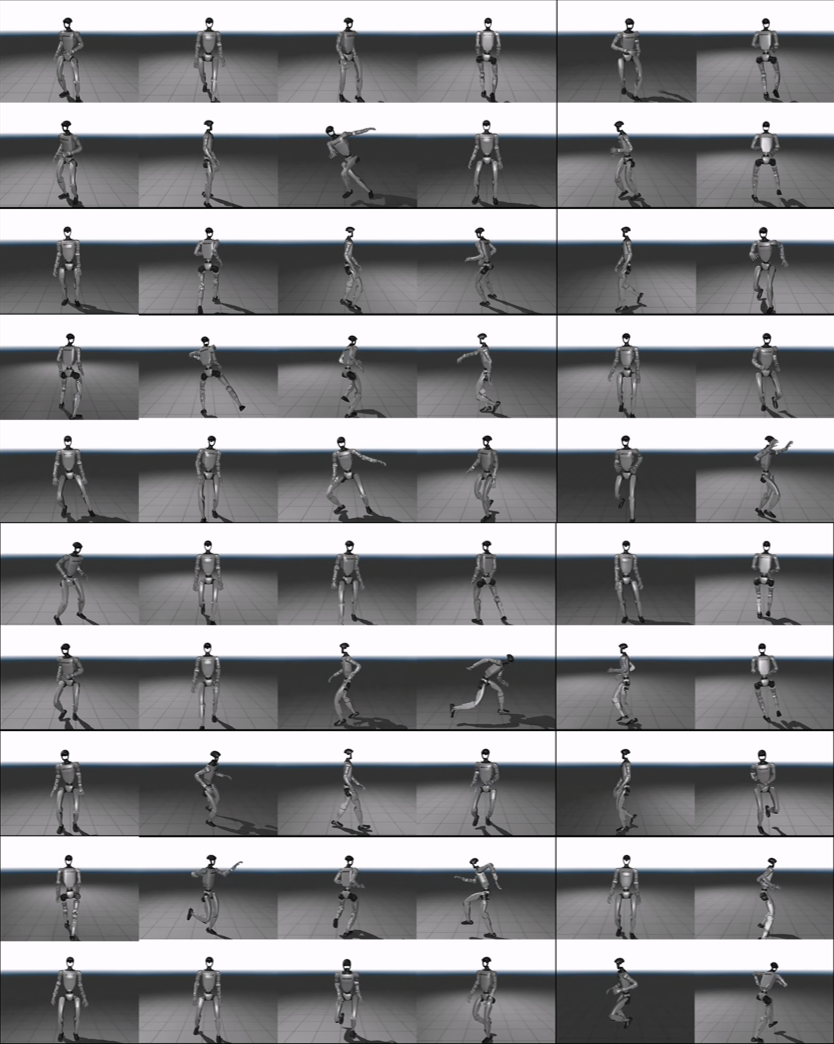}
    \caption{Visualization of IKMR refined motion results.}
    \label{fig:visresults}
\end{figure*}
\clearpage

\begin{itemize}
    \item Upper-body manipulations involving coordinated arm swinging.
    \item Highly dynamic lower-body movements, such as rapid leg kicks.
    \item Forward locomotion exhibiting a natural, human-like gait.
    \item Agile backward running characterized by a leaning-back posture.
\end{itemize}

\begin{figure}[h]
  \centering
    \includegraphics[width=1.0\linewidth]{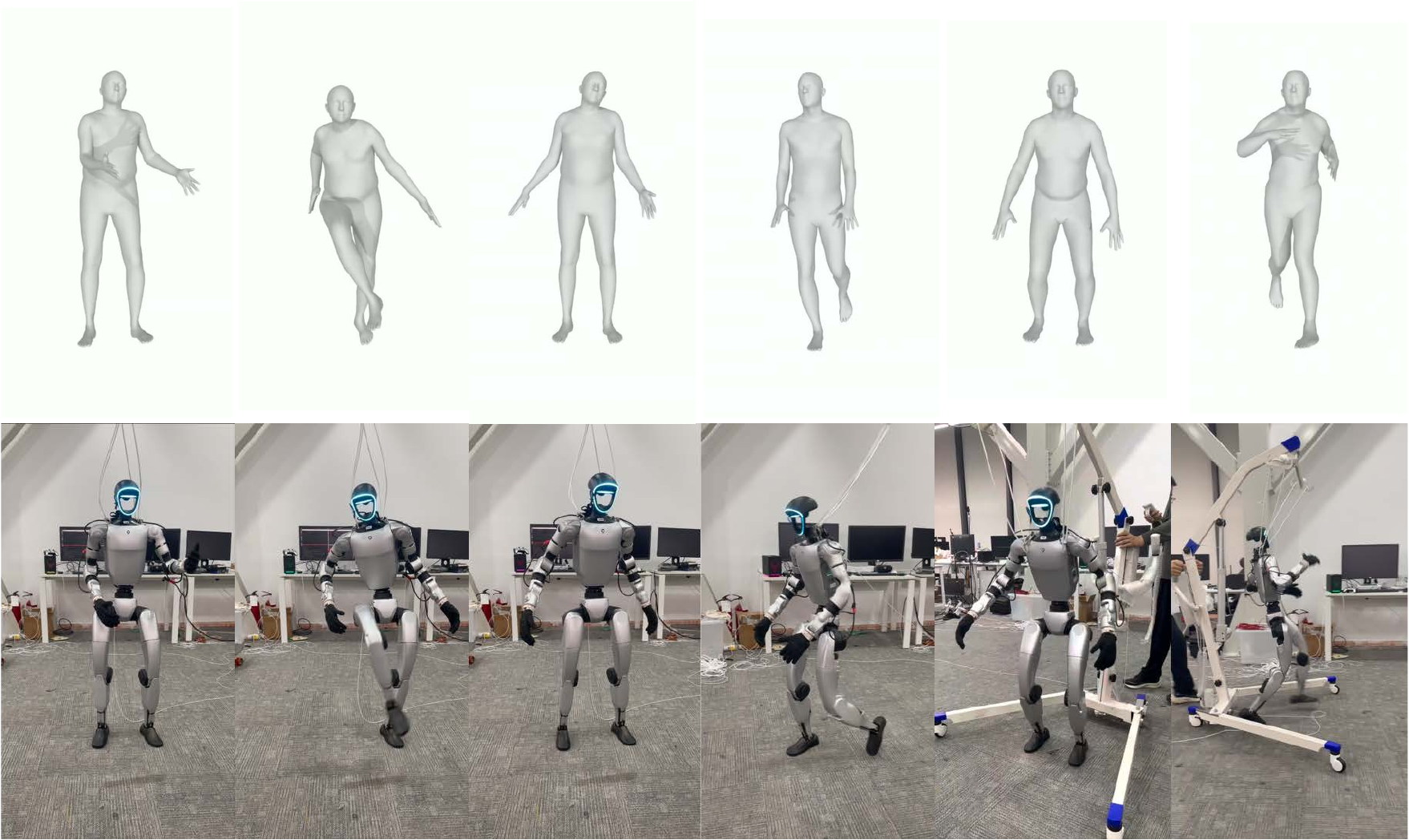}
    \caption{Real-world hardware deployment. Downstream whole-body controllers accurately reproduce the human-like stylistic nuances embedded within the IKMR-transformed trajectories.}
    \label{fig:real}
    \vspace{-6pt}
\end{figure}

Our implicit transformation pipeline enables the downstream policy to accurately capture and reproduce nuanced human-like stylistic details—such as hip rotation and sway during forward walking, and torso stabilization during backward running. We validate the reliability of our pipeline on actual devices through these representative behaviors. These real-world deployments verify that IKMR serves as a robust, end-to-end data engine, successfully converting unconstrained human motion collections into high-quality, executable training assets for physical humanoid learning.
\section{Limitation}
\label{sec:limitation}

While IKMR significantly accelerates dataset-level motion conversion, its current architectural design operates strictly on fixed-length temporal windows. Processing continuous, long-horizon in-the-wild human motions necessitates chunking the raw data into discrete sub-sequences and subsequently stitching the transformed outputs. This segmentation can introduce temporal discontinuities or subtle abrupt transitions at the boundaries of the sub-sequences, which may marginally degrade the tracking smoothness of downstream policies over extended horizons. Future iterations of this pipeline could integrate temporal attention mechanisms or autoregressive latent dynamics to ensure seamless, infinite-horizon data transformation without boundary artifacts. 

\section{Conclusion}
\label{sec:conclusion}

In this paper, we introduced Implicit Kinodynamic Motion Retargeting (IKMR), a scalable, physics-informed data transformation pipeline designed to alleviate the critical data scarcity bottleneck in humanoid robot learning. By projecting heterogeneous kinematic structures into a shared topological latent space and embedding a simulator-in-the-loop motion prior, IKMR fundamentally resolves the computational overhead and noise-sensitivity inherent in traditional frame-by-frame optimization. Extensive empirical evaluations demonstrate that our framework operates as an intrinsic data curation engine—capable of achieving high-throughput dataset conversion while automatically purifying noisy, unconstrained human motions into safe, dynamically feasible trajectories for physical hardware. 

Moving forward, expanding the shared topological representation to encompass a broader spectrum of robotic morphologies presents a highly promising research direction. Ultimately, scalable data pipelines like IKMR will serve as the indispensable bridge between multimodal generative models and physical robot learning, unlocking the vast potential of in-the-wild data to continuously fuel the next generation of humanoid foundation models.


\bibliographystyle{plainnat}
\bibliography{references}

\clearpage

\end{document}